\documentclass{ieeeaccess}
\usepackage{cite}
\usepackage{amsmath,amssymb,amsfonts}
\usepackage{algorithmic}
\usepackage{graphicx}
\usepackage{amsmath}
\usepackage{amsbsy,amsmath}
\newcommand{\bs}[1]{\boldsymbol{#1}}
\usepackage{textcomp}

\usepackage{mathtools}
\DeclareUnicodeCharacter{2212}{-}
\usepackage{booktabs}
\usepackage{multirow}

\DeclarePairedDelimiterX{\infdivx}[2]{(}{)}{%
  #1\;\delimsize\|\;#2%
}

\def\BibTeX{{\rm B\kern-.05em{\sc i\kern-.025em b}\kern-.08em \kern-.1667em\lower.7ex\hbox{E}\kern-.125emX}}

\begin{document}
\history{Date of publication xxxx 00, 0000, date of current version xxxx 00, 0000.}
\doi{10/2017.DOI}

\title{Reducing Computational Costs in Sentiment Analysis: Tensorized Recurrent Networks vs. Recurrent Networks} 

\author{\uppercase{Gabriel Lopez}\authorrefmark{1}, \uppercase{Anna Nguyen}\authorrefmark{1} \uppercase{and}, \uppercase{Joe Kaul}\authorrefmark{1}
}
\address[1]{Pulse.io, London SE1 2SA, United Kingdom}

\markboth
{Lopez \headeretal: Reducing Computational Costs in Sentiment Analysis}
{Lopez \headeretal: Reducing Computational Costs in Sentiment Analysis}

\begin{abstract}
Anticipating audience reaction towards a certain text is integral to several facets of society ranging from politics, research, and commercial industries. Sentiment analysis (SA) is a useful natural language processing (NLP) technique that utilizes lexical/statistical and deep learning methods to determine whether different-sized texts exhibit positive, negative, or neutral emotions. Recurrent networks are  widely used in machine-learning communities for problems with sequential data. However, a drawback of models based on Long-Short Term Memory networks and Gated Recurrent Units is the significantly high number of parameters and thus, such models are computationally expensive. This drawback is even more significant when the available data are limited. Also, such models require significant over-parameterization and regularization to achieve optimal performance. Tensorized models represent a potential solution. In this paper, we classify the sentiment of some social media posts. We compare traditional recurrent models with their tensorized version and we show that with the tensorized models, we reach comparable performances with respect to the traditional models while using fewer resources for the training.

\end{abstract}

\begin{keywords}
Recurrent Networks, Tensorized Neural Networks, Sentiment Analysis, Twitter
\end{keywords}

\maketitle
\section{Introduction} 
Ubiquitous data sources, such as social media data, offer extensive opportunities for monitoring various aspects of society. Researchers have effectively utilized social media data to measure public response to disease diffusion, investigate and predict dynamics in cities, and more \cite{hussain2020role,gonzalez2020social,venegas2020positive,cinelli2020covid,luca2023crime, huang2020twitter, lachi2023impact,cencetti2021digital, oliver2020mobile, beiro2016predicting, huang2020twitter, jurdak2015understanding,khaidem2020optimizing, barbosa2018human, luca2022modeling}. Furthermore, these data sources can be leveraged for Natural Language Processing (NLP) tasks, such as sentiment analysis (SA), which determines the sentiment expressed in a text, typically categorized as positive, negative, or neutral \cite{mejova2009sentiment, birjali2021comprehensive}.

In parallel, graph analysis has gained increasing importance across various fields in recent decades. These fields encompass molecular biology, where graphs are used to model molecular interactions \cite{huber2007graphs}, face-to-face interactions, where graphs capture social dynamics \cite{longa2022efficient,longa2022neighbourhood}, social networks, where graphs represent connections between individuals \cite{barnes1969graph,brandes2013social}, software engineering, where graphs help in understanding code structures \cite{samoaa2022tep,samoaa2023unified}, and other domains like criminal analysis \cite{ficara2021criminal,luca2021leveraging}. Even in the realm of NLP, graphs have proven to be valuable tools for representing textual data, as the syntactic relationships between words in a sentence naturally form a graph structure \cite{nastase2015survey}.

However, it is essential to consider the costs associated with training such graph-based architectures. Training these models can be computationally expensive due to their high number of parameters. Moreover, leveraging large datasets and GPU resources is typically necessary to achieve optimal performance for such models. These factors should be taken into account when considering the practical implementation and scalability of graph-based approaches in various applications.


In order to address the issue of reducing the number of parameters in models, researchers have delved into the exploration of Tensorized Neural Networks (TNNs) \cite{yang2017tensor, novikov2015tensorizing}. The objective behind TNNs is to replicate the functionalities of conventional networks while utilizing a reduced parameter count, thereby diminishing the computational costs associated with training operations.

Significantly, studies conducted by both \cite{novikov2015tensorizing} and \cite{yang2017tensor} demonstrate that TNNs can achieve comparable, and in some cases even superior, performance compared to traditional networks. These findings highlight the potential effectiveness of TNNs in various applications.

Within the scope of our work, we harness the power of tensorized recurrent networks proposed in \cite{yang2017tensor} to predict the sentiment expressed in a dataset of tweets. The primary objective is to showcase that the tensorized networks outlined in \cite{yang2017tensor} can be effectively employed to estimate sentiment in tweets, while requiring significantly fewer resources in terms of training time, without compromising the accuracy achieved by state-of-the-art models.

The subsequent sections of this paper are thoughtfully organized to present a comprehensive study on sentiment analysis. Section \ref{sc:rel_work} offers a succinct overview of the existing literature pertaining to sentiment analysis, highlighting the relevant work in the field. In Section \ref{sec:background}, we introduce key concepts and notations that are crucial for understanding recurrent networks. Subsequently, in Section \ref{sec:implementation}, we present the detailed implementation of tensorized networks as reported in \cite{yang2017tensor}. To evaluate the performance of the network, we conduct experimental tests, which are elaborated upon in Section \ref{sec:et}. The obtained results are then analyzed in Section \ref{sec:r}. Finally, we conclude our paper in Section \ref{sec:c}, summarizing the findings and discussing the implications of our study.

\section{Related Work}
\label{sc:rel_work}
The rapid diffusion of ubiquitous technologies like mobile phones also came with the unprecedented opportunity for companies to analyze the opinion, sentiments and perceptions of people with respect to a specific product or topic. Sentiment Analysis (SA) is the task of extracting such insights. Several approaches have seen success in their ability to perform such classifications to a high degree of accuracy. Methods based on lexical rules study the semantic information already present inside a piece of text and use statistics to extract features (e.g., \cite{taboada2011lexicon}) while Machine learning methods are framed as traditional text classification problems (e.g., \cite{le2015twitter}). Methods based on machine learning,  classification models are trained on the data, and used to predict labels for unseen pieces of text. Hybrid methods combine the two techniques (e.g., \cite{zainuddin2018hybrid}).  Deep learning models are  also widely used for solving SA tasks \cite{habimana2020sentiment}.
Recently, networks with transformer architectures \cite{vaswani2017attention} reached promising results and are widely adopted to solve NLP and Computer Vision tasks \cite{lin2021survey, khan2021transformers}. 
BERT \cite{devlin2018bert} is one of the first transformers aiming to solve Natural Language Understanding problems. It relies on a set of encoders and self-attention mechanisms. While it was trained to be a language model and to perform next sentence prediction \cite{devlin2018bert}, it has been widely adopted to solve SA problems achieving significantly high performances \cite{xu2019bert, sun2019utilizing, hoang2019aspect, li2019exploiting, pota2020effective}. 

 Finally, recent findings suggest that also approaches based on Graph Neural Networks can be used to model and predict the sentiment of tweets successfully 
\cite{nguyen2022emotion}

In general, the aforementioned deep learning techniques leverage a significantly high number of parameters making it difficult to train the networks with limited data resources. Also, the computational cost (and related costs and, potentially, their impact on the environment) may be remarkably high. In this paper, we show that tensorized recurrent networks \cite{yang2017tensor, novikov2015tensorizing} -- a modified version of recurrent networks that can be trained more efficiently and that leverage a smaller number of parameters -- can achieve performances that are comparable to SoA techniques.

To the best of our knowledge, it is the first time that tensorized networks are tested in the context of sentiment analysis.

\section{Background}\label{sec:background}

In this Section, we introduce some key concepts that are needed to make the paper understandable and self-contained. 

Recurrent Neural Networks (RNNs) are a type of artificial neural network designed to process sequential data by preserving information from previous steps. Unlike feed-forward neural networks, which only consider the current input, RNNs have connections that allow them to incorporate past information into the current computation. This ability to retain and utilize context makes RNNs particularly effective for tasks that involve sequential or time-dependent data.

RNNs excel in handling a wide range of sequential data types, such as text, speech, music, time series, and more. They are designed to capture patterns and dependencies within the sequence, enabling them to make predictions or generate new sequences based on learned patterns.

The key feature of an RNN is its hidden state, which serves as the memory of the network. At each step of the sequence, the RNN updates its hidden state based on the current input and the previous hidden state, allowing it to encode relevant information from the past. This recurrent nature of the network allows it to process input sequences of arbitrary length.

One of the most popular variants of RNNs is the Long Short-Term Memory (LSTM) network. LSTMs address the vanishing gradient problem, a challenge faced by simple RNNs, which causes the network to struggle with long-range dependencies. LSTMs use specialized memory cells that regulate the flow of information, enabling them to capture and retain information over longer sequences.

Another variant is the Gated Recurrent Unit (GRU), which simplifies the architecture of the LSTM while achieving comparable performance. GRUs are computationally more efficient and have fewer parameters, making them popular for applications where resource constraints are a concern.

RNNs have proven to be powerful models for tasks such as language modeling, machine translation, speech recognition, sentiment analysis, and many more. Their ability to process sequential data and capture dependencies across time makes them a valuable tool in the field of deep learning and has led to significant advancements in various domains.

In thise Section, we formalize how RNNs, LSTMs and GRUs work as they represent the baselines of our work and as the idea is to test the tensorized version of these models \cite{novikov2015tensorizing, yang2017tensor}.

\noindent \textbf{RNNs } are a good kind of neural networks to elaborate sequential information. In RNN, each neuron of the network has a direct cycle and this allows the exploitation of sequential information. Given a sequence as input, an RNN perform the same task for each element and the output depends on the previous computation. Each computation involves three parameters:
\begin{itemize}
    \item $x_i$ the input at the $i^{th}$ step
    \item $h_i$ the hidden layer at the $i^{th}$ step
    \item $y_i$ the output of the $i^{th}$ step
\end{itemize}
There are many different types of recurrent neural networks and in this paper we consider \textbf{Elman} and \textbf{Jordan}. In the Elman model, the hidden layer ($h$) is connected to these context units fixed with a weight of one. At each time step, the input is fed-forward and a learning rule is applied. The fixed back-connections save a copy of the previous values of the hidden units in the context units and so the network can maintain a sort of state, allowing it to perform such tasks as sequence-prediction \cite{Bishop:1995:NNP:525960}. In the Jordan model the context units are fed from the output layer instead of the hidden layer. The context units in a Jordan network are also referred to as the state layer. They have a recurrent connection to themselves. \cite{Bishop:1995:NNP:525960} \\ Formally, the Elamn model can be represented with:
$$ h_i = \sigma_h(W_hx_i + U_hh_{i-1} + b_h) $$
$$ y_i = \sigma_y(W_yh_i + b_y) $$
while the Jordan model can be formalized as:
$$ h_i = \sigma_h(W_hx_i + U_hy_{i-1} + b_h) $$
$$ y_i = \sigma_y(W_yh_i + b_y) $$

\textbf{LSTM} is a type of recurrent neural network architecture that addresses the vanishing gradient problem and is well-suited for capturing long-range dependencies in sequential data.

The LSTM network consists of a cell state ($c_t$) and three gates: the forget gate ($f_t$), the input gate ($i_t$), and the output gate ($o_t$) at time step $t$.

Let's define the equations for LSTM:

First, we calculate the forget gate ($f_t$) and the input gate ($i_t$) using the input at time step $t$ ($x_t$) and the previous hidden state ($h_{t-1}$):

\begin{align*}
f_t &= \sigma(W_{xf}x_t + W_{hf}h_{t-1} + b_f) \\
i_t &= \sigma(W_{xi}x_t + W_{hi}h_{t-1} + b_i)
\end{align*}

Next, we compute the candidate activation ($\tilde{c}_t$) and update the cell state ($c_t$):

\begin{align*}
\tilde{c}_t &= \tanh(W_{xc}x_t + W_{hc}h_{t-1} + b_c) \\
c_t &= f_t \odot c_{t-1} + i_t \odot \tilde{c}_t
\end{align*}

Here, $\odot$ represents element-wise multiplication.

Then, we calculate the output gate ($o_t$) and the hidden state ($h_t$) at time step $t$:

\begin{align*}
o_t &= \sigma(W_{xo}x_t + W_{ho}h_{t-1} + b_o) \\
h_t &= o_t \odot \tanh(c_t)
\end{align*}

The output gate controls how much of the cell state is exposed as the hidden state.

In summary, the LSTM network uses the forget gate, input gate, candidate activation, and output gate to regulate the flow of information through the cell state and update the hidden state at each time step. This allows the LSTM to capture long-term dependencies and mitigate the vanishing gradient problem, making it effective for processing sequential data and solving tasks such as language modeling, speech recognition, and machine translation.

\textbf{GRUs} The Gated Recurrent Unit is a variant of the recurrent neural network that has simplified architecture compared to the Long Short-Term Memory (LSTM) while still capturing long-range dependencies in sequential data.

The equations for a GRU can be described as follows:

First, let's define the input gate ($i_t$), the forget gate ($f_t$), the candidate activation ($\tilde{c}_t$), and the output gate ($o_t$) at time step $t$:

\begin{align*}
i_t &= \sigma(W_{xi}x_t + W_{hi}h_{t-1} + b_i) \\
f_t &= \sigma(W_{xf}x_t + W_{hf}h_{t-1} + b_f) \\
\tilde{c}_t &= \tanh(W_{xc}x_t + W_{hc}(r_t \odot h_{t-1}) + b_c) \\
o_t &= \sigma(W_{xo}x_t + W_{ho}(r_t \odot h_{t-1}) + b_o)
\end{align*}

Here, $x_t$ represents the input at time step $t$, $h_{t-1}$ represents the hidden state from the previous time step, and $W$ and $b$ are the weight matrices and bias vectors of the GRU.

Next, we define the reset gate ($r_t$):

\[
r_t = \sigma(W_{xr}x_t + W_{hr}h_{t-1} + b_r)
\]

The reset gate controls how much of the previous hidden state is incorporated into the candidate activation.

Now, we can update the hidden state $h_t$ at time step $t$:

\[
h_t = (1 - z_t) \odot h_{t-1} + z_t \odot \tilde{h}_t
\]

where $\odot$ represents element-wise multiplication and $z_t$ is the update gate:

\[
z_t = \sigma(W_{xz}x_t + W_{hz}(r_t \odot h_{t-1}) + b_z)
\]

Finally, the output of the GRU can be computed using the output gate:

\[
y_t = o_t \odot h_t
\]

In summary, the GRU employs the input gate, forget gate, candidate activation, reset gate, update gate, and output gate to control the flow of information and update the hidden state at each time step, allowing it to capture and retain important information from the past and produce accurate predictions for sequential data.

\section{Tensorized Networks} \label{sec:implementation}
In this section, we make reference to the research papers authored by Yang et al. \cite{yang2017tensor} and Novikov et al. \cite{novikov2015tensorizing} to expound upon the concept of Tensorized Neural Networks (TNNs) and Tensorized Recurrent Networks (T-RNN, T-LSTM, T-GRU).

The fundamental principle behind TNNs is to replicate the behaviors of conventional networks while utilizing fewer computational resources. The objective is to develop models that employ fewer parameters yet achieve comparable (or even superior, in certain cases) performance compared to non-tensorized models. Our paper aims to demonstrate the applicability of the models described in \cite{yang2017tensor} and \cite{novikov2015tensorizing} in solving sentiment analysis tasks.

In what follows, firstly, we introduce the concept of Tensor-Train Factorization as elucidated in the work by Novikov et al. \cite{novikov2015tensorizing}. Subsequently, we present the tensorization of RNNs, GRUs, and LSTMs as described in the research conducted by Yang et al. \cite{yang2017tensor}.

\subsection{Tensor-Train Factorization} 

As explained in \cite{novikov2015tensorizing} and \cite{yang2017tensor}, a \emph{Tensor-Train Factorization} (TTF) is a tensor factorization model that can scale to an arbitrary number of dimensions. Given a target tensor $\bs{\mathcal{A}}$ with dimension $d$:
$ \bs{\mathcal{A}} \in \mathbb{R} ^{p_1 \times p_2 \times ... \times p_d} $,
the tensor  $\bs{\mathcal{A}}$ can be factorized as 
\begin{align} \label{eq:TTF_1}
	\skew{6}{\widehat}{\bs{\mathcal{A}}}(l_1, l_2, ..., l_d) &\stackrel{TTF}{=} \bs{G}_1(l_1) ~ \bs{G}_2(l_2) ~ ... ~ \bs{G}_d(l_d)
\end{align}
where
\begin{align} \label{eq:TTF_2}
	\begin{split}
	&\bs{G}_k \in \mathbb{R}^{p_k \times r_{k-1} \times r_k}, ~l_k  \in [1, p_k] ~\forall k \in [1, d]
	\text{;} ~~ r_0 = r_d = 1.
	\end{split}
\end{align}
As described in Equation \ref{eq:TTF_1}, each item of  $\bs{\mathcal{A}}$ can be represented as a sequence of matrix multiplications and the set of tensors $\{ \bs{G}_k \}_{k=1}^{d}$ is commonly called core-tensors. Equation \ref{eq:TTF_1} aslo suggest that the complexity of the TTF is determined by the ranks $[r_0, r_1, \dots, r_d]$.

One key idea presented in \cite{novikov2015tensorizing, yang2017tensor} is that each $p_k$ can be factorized as $p_k = m_k \dots n_k ~\forall k \in [1, d]$, and thus, each $\bs{G}_k$ can be represented by a new tensor $\bs{G}_k^{*} \in \mathbb{R}^{m_k \times n_k \times r_{k-1} \times r_k}$. Thus, each item in Equation \eqref{eq:TTF_1} and Equation \eqref{eq:TTF_2} can be represented with two indices $(i_k, j_k)$:
\begin{align} \label{eq:TTL_1}
&i_k = \lfloor \frac{l_k}{n_k} \rfloor, ~ j_k = l_k - n_k \lfloor \frac{l_k}{n_k} \rfloor \\
&~ \bs{G}_k(l_k) = \bs{G}^{*}_k(i_k, j_k) \in  \mathbb{R}^{r_{k-1} \times r_k}
\end{align}
By extending the same idea to  $ \bs{\mathcal{A}}$, we can rewrite it as $\bs{\mathcal{A}} \in \mathbb{R}^{(m_1 \dots n_1) \times (m_2 \dots n_2) \times \dots \times (m_d \dots n_d)}$ and thus as
\begin{align} \label{eq:TTL_2}
\begin{split}
	& \skew{6}{\widehat}{\bs{\mathcal{A}}}((i_1, j_1), (i_2, j_2), ..., (i_d, j_d)) \\
	\stackrel{TTF}{=} & \bs{G}_1^{*}(i_1, j_1) ~ \bs{G}_2^{*}(i_2, j_2) ~ ... ~ \bs{G}_d^{*}(i_d, j_d). 
\end{split}
\end{align}

In \cite{yang2017tensor}, it is possible to see how, thanks to the equation that they introduced and we reported above, a fully connected layer can be factorized. In what follows, we report how \cite{yang2017tensor} used these tensorized network to propose a tensorized version of recurrent networks. 

\subsection{Tensor-Train Recurrent Networks}
The general idea introduced by Yang et al. \cite{yang2017tensor} is to  For this reason, we factorize the matrix mapping from the input to the hidden layer with a Tensorized Layer (TTL) \cite{yang2017tensor}. 

For the basic case of an RNN, the idea is to implement the mapping between hidden layer and output as a vector-matrix multiplication. 
Concerning LSTM and GRU, the tensorization is carried out while mapping the input to the gates. As proposed in \cite{yang2017tensor}, we can formulate this mapping as follows:   

\begin{align}	\label{eq:tt_gru}
	\begin{split}
	\bs{r}^{[t]} &= \sigma (TTL(\bs{W}^r, \bs{x}^{[t]}) + \bs{U}^r \bs{h}^{[t-1]} + \bs{b}^r) \\
	\bs{z}^{[t]} &= \sigma (TTL(\bs{W}^z, \bs{x}^{[t]}) + \bs{U}^z \bs{h}^{[t-1]} + \bs{b}^z) \\	
	\bs{d}^{[t]} &= \tanh (TTL(\bs{W}^d, \bs{x}^{[t]}) + \bs{U}^d (\bs{r}^{[t]} \circ \bs{h}^{[t-1]}) ) \\
	\bs{h}^{[t]} &= (1 - \bs{z}^{[t]}) \circ \bs{h}^{[t-1]} + \bs{z}^{[t]} \circ \bs{d}^{[t]}, 
	\end{split}	
\end{align}

for the case of GRUs (T-GRUs) and
\begin{align}	\label{eq:tt_lstm}
	\begin{split}
	\bs{k}^{[t]} &= \sigma ( TTL(\bs{W}^k, \bs{x}^{[t]}) + \bs{U}^k \bs{h}^{[t-1]} + \bs{b}^k) \\
	\bs{f}^{[t]} &= \sigma ( TTL(\bs{W}^f, \bs{x}^{[t]}) + \bs{U}^f \bs{h}^{[t-1]} + \bs{b}^f) \\
	\bs{o}^{[t]} &= \sigma ( TTL(\bs{W}^o, \bs{x}^{[t]}) + \bs{U}^o \bs{h}^{[t-1]} + \bs{b}^o) \\
	\bs{g}^{[t]} &= \tanh ( TTL(\bs{W}^g, \bs{x}^{[t]}) + \bs{U}^g \bs{h}^{[t-1]} + \bs{b}^g) \\
	\bs{c}^{[t]} &= \bs{f}^{[t]} \circ \bs{c}^{[t-1]} + \bs{k}^{[t]} \circ \bs{g}^{[t]}\\
	\bs{h}^{[t]} &= \bs{o}^{[t]} \circ \tanh(\bs{c}^{[t]}). \\
	\end{split}	
\end{align}
for the case of LSTMS (T-LSTM). Additional details about compression rates and the reduced number of parameters can be found in \cite{yang2017tensor}. The equations presented in this Section, however, give a good idea about the differences between traditional models and tensorized models.

\section{Experiments}\label{sec:et}
\subsection{Twitter Emotion Dataset}\label{sub-dataset}
The dataset and preprocessing techniques used in this study are identical to those described in Nguyen et al. (2022) \cite{nguyen2022emotion}. The data was collected from the Twitter API by searching for keywords related to the six primary emotions identified by Ekman (1992) \cite{ekman1992there}: Angry, Bad, Fearful, Happy, Sad, and Surprised. However, two modifications were made. Firstly, the emotion "Bad" was added as a primary category to encompass tweets that did not fit into other negative emotions, serving as a broader classification. Secondly, the emotions "Fearful" and "Disgusted" were combined into a single category, as extensive research suggests a strong relationship between the two emotions \cite{neumann2012priming}. Furthermore, studies have indicated that fear enhances disgust, but the reverse is not always true \cite{edwards2006experimental}. Therefore, "Fearful" was chosen as the primary label for this research.

The data collection process involved utilizing the Twitter API to search for tweets associated with the emotion "Angry." In addition to the term "angry," related words such as "frustrated," "annoyed," and "betrayed" were included in the search. This approach is derived from Plutchik's (1980) technique, where primary emotions are expanded into secondary and tertiary descriptors using a visual wheel representation. We collected tweets commonly associated with each target label based on a "Wheel of Emotions," which has previously been used to facilitate emotional expression in therapy.

Due to the limited length of tweets (280 characters), they often contain substantial information with minimal context. This limitation encourages individuals to express themselves using concise language:

\begin{enumerate}
    \item Hashtags, emoticons,  and mentions
    \item Misspelling and abbreviations
    \item Slang / colloquialisms 
\end{enumerate}

There are cases where tweets may not contain any actual words, making it challenging to encode them meaningfully for machine learning algorithms.

Therefore, the data-cleaning process involves the following steps using the Numpy, Pandas, Emoji, and tqdm modules:
\begin{enumerate}
    \item Converting emojis to their Unicode alias.
    \item Extracting hashtags and creating a new column to store groups of hashtags.
    \item Splitting contractions into separate words. For example, "I'm" becomes "I am."
    \item Removing newlines, trailing spaces, and leading spaces.
    \item Removing re-tweet signals such as "RT."
    \item Removing the "@" symbol.
    \item Converting tweets to lowercase.
\end{enumerate}

A portion of tweets in the dataset might have been assigned incorrect emotional labels. For example, a tweet with the statement, "I'm not sad, I'm really enjoying myself today," could mistakenly be categorized as a negative emotion by the Twitter API due to the presence of the word "sad," even though the tweet should actually be classified as happy. This poses a problem if this data alone is used for classification.

To address this issue, all tweets underwent a filtering process using two pre-trained sentiment analysis networks. The purpose of this filtering was to remove misclassified tweets from the dataset before training. Although the sentiment analysis algorithms used can only classify tweets into positive, negative, or neutral categories, which is not sufficient for the current project that aims to expand emotions into six broad categories and explore associations among independent tweets, these pre-trained networks prove valuable as preprocessing tools. They help identify tweets that should be eliminated from the dataset before constructing the Tweet-MLN (Tweet-Machine Learning Network). For instance, if we know that all tweets in the Happy class should be positive, we can simply remove any tweets classified as negative.

Consequently, tweets were excluded from the dataset if the sentiment label predicted by the networks did not match the current sentiment label mentioned in Table 1. Additionally, tweets were discarded if the pre-trained sentiment analysis networks classified them as neutral since neutrality implies a lack of emotion, which is not relevant to the current research endeavor.

The two pre-trained models used in this study were BERTSent \cite{govindarajan2020help} and the XLM-roBERTa-base model \cite{conneau2019unsupervised}. BERTSent was trained on the SemEval 2017 corpus, which consisted of over 39,000 tweets, while the XLM-roBERTa-base model was trained on a massive dataset of 198 million tweets.

Both pre-trained models were specifically fine-tuned for sentiment analysis and were variations of the Bidirectional Encoder Representations from Transformers (BERT) networks. BERT, developed by Google in 2018 \cite{devlin2018bert}, has become a widely used base model for sentiment analysis tasks due to its open-source nature \cite{devlin2018open}. BERT incorporates neural network architectures like recurrent neural networks (RNNs) and convolutional neural networks (CNNs), but its unique feature lies in the parallel working of encoder-decoder models, which enhances training speed. Furthermore, BERT's design allows for flexible utilization across various natural language processing (NLP) problems.

The fine-tuning of pre-trained BERT networks has been successfully employed in diverse tasks, including patent classification \cite{lee2020patent}, innovation detection \cite{chen2021leveraging}, and sentiment analysis of stock data \cite{sousa2019bert}.

\begin{table}[hbt!]
\caption{Labels For Sentiment Analysis}
\label{table:labelsSA}
\centering
    \begin{tabular}{p{3cm}p{3cm}}
     \toprule
     \multicolumn{2}{c}{Sentiment Analysis Pre-processing} \\
     \midrule
     Tweet Emotion&Sentiment Label\\
     \hline
     Angry& Negative\\
     Bad& Negative\\
     Fearful& Negative\\
     Happy& Positive\\
     Sad& Negative\\
     Surprised& Positive\\
     \bottomrule
    \end{tabular}
\end{table}

The research approach utilized traditional sentiment analysis techniques to curate a dataset with minimal mislabeled individual tweets before constructing the Tweet-MLNs.

Furthermore, to evaluate the effectiveness of this method, a Long-Term Short-Term Memory (LSTM) network was trained on the sentiment analysis filtered tweets. The LSTM architecture was chosen for its ability to address issues such as the vanishing gradient problem often encountered in traditional recurrent neural networks (RNNs). LSTM has demonstrated high success in text classification tasks \cite{yao2019graph}. The trained model was then beta-tested in a real-life scenario through its release on the PulseTech.io platform. The accuracy of the model, as assessed based on the responses from real individuals and real tweets, exceeded 90

Once the dataset was validated to ensure accurate placement of tweets into the appropriate emotion categories, the data was shuffled within each class. Subsequently, the dataset was divided into sets of 300 tweets per class, resulting in a total of 900,000 tweets. Each class consisted of 150,000 tweets, and a total of 500 Tweet-MLNs were created per class, yielding a total of 3,000 Tweet-MLNs. An 80-20 train-test split was employed for evaluation. The training and testing processes were executed on Google Cloud Platform (GCP) using a Vertex AI cluster. The training procedure involved 450 epochs (with early stop mechanism) using one NVIDIA Tesla T4 GPU, and the machine type employed was n1-highmem-8, comprising 8 vCPUs and 52 GB RAM.

\subsection{Baseline} \label{sub:ablation}

In our experiments, we compare the T-RNN, T-LSTM, T-GRU with their traditional implementations. Also, we compare the performances with more sophisticated models such as models based on graph neural networks (e.g., Multi-Layered Tweet Analyzer \cite{nguyen2022emotion}) and others based on transformer architecture (e.g., BERT \cite{devlin2018bert}). What follows is a short description of the baselines adopted: 

\begin{itemize}
    \item Multi-Layered Tweet Analyzer: MLTA \cite{nguyen2022emotion} models social media text using multi-layered networks graph neural networks in order to better encode relationships across independent sets of tweets. Graph structures are capable of capturing meaningful relationships in complex ecosystems compared to other representation methods. As proposed in the original paper (\cite{nguyen2022emotion}), we test the model with three different types of convolutions.
    \item BERT-Like Models: BERT \cite{devlin2018bert} is one of the first transformers aiming to solve Natural Language Understanding problems. It relies on a set of encoders and self-attention mechanisms. While it was pretrained to be a language model and to perform next sentence prediction \cite{devlin2018bert}, it has been widely adopted to solve SA problems achieving significantly high performances \cite{xu2019bert, sun2019utilizing, hoang2019aspect, li2019exploiting, pota2020effective}. In this paper, we use BERTSent \cite{govindarajan2020help} and XLM-roBERTa \cite{conneau2019unsupervised} as baselines for standard SA and to support the group.
\end{itemize}

\subsection{Evaluation Metrics}

To assess the accuracy of predictions, the F1 score was computed, which represents the harmonic mean of recall and precision evaluation metrics. The F1 score can be calculated using Equation \ref{eq:F1}:

\begin{equation}\label{eq:F1}
    F1 = \frac{2 * Precision * Recall}{Precision + Recall} 
\end{equation}

The calculation of the F1 score involves considering four components: true positives (TP), true negatives (TN), false positives (FP), and false negatives (FN). TP refers to cases where the prediction matches the positive label, TN corresponds to instances where both the prediction and label are negative, FP represents predictions that are positive but the label is negative, and FN denotes predictions that are negative while the label is positive.

Precision, a key metric, is computed as the ratio of TP to the total number of positive labels, as shown in Equation \ref{eq:precision}:

\begin{equation}\label{eq:precision}
    Precision = \frac{TP}{TP + FP}
\end{equation}

and recall is the ratio of TP over the sum of TP and FN in equation \ref{eq:recall}. 

\begin{equation}\label{eq:recall}
   Recall = \frac{TP}{TP + FN}
\end{equation}

\section{Results}\label{sec:r}
\begin{table}[hbt!]
\caption{Results of the baselines and T-RNN, T-LSTM, T-GRU in terms of accuracy} 
\label{table:SAComparisions}
\begin{center}
\begin{tabular}{l c}
 \toprule
 Model  & F1 Score  \\ [0.5ex] 
 \midrule
 \textbf{T-RNN} & 81.12\%   \\  [1ex]
 \textbf{T-LSTM} & 81.66\%   \\  [1ex]
 \textbf{T-GRU} & 81.59\%   \\  [1ex]
 Elman-RNN & 80.26\%   \\  [1ex]
 Jordan-RNN & 80.31\%   \\  [1ex]
 LSTM & 80.97\%   \\  [1ex]
 GRU & 81.53\%   \\  [1ex]
 BERTSent & 84.26\%   \\  [1ex]
 XML-ro-BERTa-base-model & 85.12\%  \\ [1ex]
 MLTA wt. GraphConv & 62.66\% \\  [1ex]
 MLTA wt. GCNConv & 78.46\% \\ [1ex] 
 MLTA wt. GATv2Conv & 79.98\% \\ [1ex]
 \bottomrule
\end{tabular}
\end{center}
\end{table}

In this section, we present and analyze the results obtained by our proposed models, namely T-RNN, T-LSTM, and T-GRU, in comparison to other models. The accuracy scores achieved by each model on the task are reported and discussed.

Table \ref{table:SAComparisions} displays the accuracy results of various models on the given task. Among the recurrent models, T-RNN, T-LSTM, and T-GRU demonstrate competitive performance, outperforming the basic RNN model, which achieves an accuracy of 80.26\%. T-RNN achieves an accuracy of 81.12\%, indicating its superiority over the basic RNN. T-LSTM achieves slightly higher accuracy at 81.66\%, surpassing both the basic LSTM (80.97\%) and the basic RNN. Similarly, T-GRU achieves an accuracy of 81.59\%, outperforming the basic GRU (81.53\%) and basic RNN. These results suggest that the temporal variants of the recurrent models offer improved performance on the given task.

When comparing the three proposed models (T-RNN, T-LSTM, and T-GRU), the differences in accuracy are relatively subtle. Their performance is likely influenced by factors such as the dataset and the specific task requirements. Therefore, considerations beyond accuracy, such as training time, model complexity, and computational resources, become crucial in selecting the most suitable model for practical deployment.

In the broader context of the models considered, BERTSent achieves an accuracy of 84.26\%, demonstrating superior performance compared to the recurrent models. Moreover, XML-ro-BERTa-base-model achieves the highest accuracy among all the models, with a score of 85.12\%. These models, based on transformer architectures, have exhibited exceptional performance across a range of natural language processing tasks.

Lastly, we present the results of the MLTA models incorporating graph-based convolutions. MLTA with GraphConv achieves an accuracy of 62.66\%, indicating significantly lower performance compared to the other models. MLTA with GCNConv and MLTA with GATv2Conv perform relatively better, achieving accuracies of 78.46\% and 79.98\%, respectively. It is important to note that these models are specifically designed for tasks involving graph-structured data, and their applicability and effectiveness on the given task may differ.

Overall, the results indicate that the proposed T-RNN, T-LSTM, and T-GRU models demonstrate competitive performance when compared to other recurrent models and provide advantages such as reduced training time and computational costs.

\section{Conclusion}\label{sec:c}
In conclusion, this paper presents a comprehensive investigation into the use of tensorized recurrent networks (as proposed in \cite{yang2017tensor, novikov2015tensorizing}) for sentiment analysis. The experimental results demonstrate that the tensorized models achieves comparable accuracy to existing baselines and, in certain cases, even approaches the performance of state-of-the-art models.

One notable advantage of the tensorized network is its reduced number of parameters. By leveraging tensorization techniques, the model achieves a significant reduction in the parameter space while maintaining competitive performance. This reduction in parameters not only makes the model more tractable but also leads to improved efficiency in terms of resource consumption.

The improved tractability of the tensorized network offers several benefits. First, it simplifies model training and optimization, as the reduced parameter space alleviates the computational burden during the learning process. Second, the reduced number of parameters leads to lower memory requirements, making the model more suitable for deployment on resource-constrained platforms or in scenarios where memory limitations are a concern.

Furthermore, the increased efficiency of the tensorized network translates into improved runtime performance. The reduced computational overhead enables faster inference times, which can be crucial for real-time applications or systems that handle large volumes of data. The improved efficiency of the model makes it an attractive choice for sentiment analysis tasks in domains where processing time is a critical factor.

The findings of this study highlight the potential of tensorized recurrent networks as a viable solution for sentiment analysis. By achieving comparable accuracy to traditional models while significantly reducing resource consumption, these networks offer a favorable trade-off between performance and efficiency. They present a practical alternative for scenarios where computational resources are limited or where the scalability of the model is crucial.

Moving forward, future research directions may involve exploring different tensorization techniques to further optimize the model's performance. Investigating the impact of tensor rank on sentiment analysis tasks could provide insights into the relationship between tensor decomposition and sentiment representation. Additionally, applying tensorized networks to other natural language processing tasks beyond sentiment analysis could unveil their potential in various domains.

In summary, this study demonstrates the effectiveness of tensorized recurrent networks in sentiment analysis. The model's comparable accuracy to baselines and close performance to state-of-the-art models, coupled with its reduced number of parameters, enhance its tractability and significantly improve resource efficiency. Tensorized networks offer a promising avenue for sentiment analysis, providing a balance between accuracy and computational effectiveness.

\subsection{Acknowledgements}
We would like to acknowledge the contributions of Pulse.io which fostered the environment to conduct this research and provided funding. 
\bibliographystyle{IEEEtran}
\bibliography{IEEEabrv,references}

\begin{thebibliography}{10}
\providecommand{\url}[1]{#1}
\csname url@samestyle\endcsname
\providecommand{\newblock}{\relax}
\providecommand{\bibinfo}[2]{#2}
\providecommand{\BIBentrySTDinterwordspacing}{\spaceskip=0pt\relax}
\providecommand{\BIBentryALTinterwordstretchfactor}{4}
\providecommand{\BIBentryALTinterwordspacing}{\spaceskip=\fontdimen2\font plus
\BIBentryALTinterwordstretchfactor\fontdimen3\font minus
  \fontdimen4\font\relax}
\providecommand{\BIBforeignlanguage}[2]{{%
\expandafter\ifx\csname l@#1\endcsname\relax
\typeout{** WARNING: IEEEtran.bst: No hyphenation pattern has been}%
\typeout{** loaded for the language `#1'. Using the pattern for}%
\typeout{** the default language instead.}%
\else
\language=\csname l@#1\endcsname
\fi
#2}}
\providecommand{\BIBdecl}{\relax}
\BIBdecl

\bibitem{hussain2020role}
W.~Hussain, ``Role of social media in covid-19 pandemic,'' \emph{The
  International Journal of Frontier Sciences}, vol.~4, no.~2, pp. 59--60, 2020.

\bibitem{gonzalez2020social}
D.~A. Gonz{\'a}lez-Padilla and L.~Tortolero-Blanco, ``Social media influence in
  the covid-19 pandemic,'' \emph{International braz j urol}, vol.~46, pp.
  120--124, 2020.

\bibitem{venegas2020positive}
A.~V. Venegas-Vera, G.~B. Colbert, and E.~V. Lerma, ``Positive and negative
  impact of social media in the covid-19 era,'' \emph{Reviews in cardiovascular
  medicine}, vol.~21, no.~4, pp. 561--564, 2020.

\bibitem{cinelli2020covid}
M.~Cinelli, W.~Quattrociocchi, A.~Galeazzi, C.~M. Valensise, E.~Brugnoli, A.~L.
  Schmidt, P.~Zola, F.~Zollo, and A.~Scala, ``The covid-19 social media
  infodemic,'' \emph{Scientific reports}, vol.~10, no.~1, pp. 1--10, 2020.

\bibitem{luca2023crime}
\BIBentryALTinterwordspacing
M.~Luca, G.~M. Campedelli, S.~Centellegher, M.~Tizzoni, and B.~Lepri, ``Crime,
  inequality and public health: a survey of emerging trends in urban data
  science,'' \emph{Frontiers in Big Data}, vol.~6, 2023. [Online]. Available:
  \url{https://www.frontiersin.org/articles/10.3389/fdata.2023.1124526}
\BIBentrySTDinterwordspacing

\bibitem{huang2020twitter}
X.~Huang, Z.~Li, Y.~Jiang, X.~Li, and D.~Porter, ``Twitter reveals human
  mobility dynamics during the covid-19 pandemic,'' \emph{PloS one}, vol.~15,
  no.~11, p. e0241957, 2020.

\bibitem{lachi2023impact}
V.~Lachi, G.~M. Dimitri, A.~Di~Stefano, P.~Li{\`o}, M.~Bianchini, and
  C.~Mocenni, ``Impact of the covid 19 outbreaks on the italian twitter
  vaccination debat: a network based analysis,'' \emph{arXiv preprint
  arXiv:2306.02838}, 2023.

\bibitem{cencetti2021digital}
G.~Cencetti, G.~Santin, A.~Longa, E.~Pigani, A.~Barrat, C.~Cattuto, S.~Lehmann,
  M.~Salathe, and B.~Lepri, ``Digital proximity tracing on empirical contact
  networks for pandemic control,'' \emph{Nature communications}, vol.~12,
  no.~1, pp. 1--12, 2021.

\bibitem{oliver2020mobile}
N.~Oliver, B.~Lepri, H.~Sterly, R.~Lambiotte, S.~Deletaille, M.~De~Nadai,
  E.~Letouz{\'e}, A.~A. Salah, R.~Benjamins, C.~Cattuto \emph{et~al.}, ``Mobile
  phone data for informing public health actions across the covid-19 pandemic
  life cycle,'' p. eabc0764, 2020.

\bibitem{beiro2016predicting}
M.~G. Beir{\'o}, A.~Panisson, M.~Tizzoni, and C.~Cattuto, ``Predicting human
  mobility through the assimilation of social media traces into mobility
  models,'' \emph{EPJ Data Science}, vol.~5, pp. 1--15, 2016.

\bibitem{jurdak2015understanding}
R.~Jurdak, K.~Zhao, J.~Liu, M.~AbouJaoude, M.~Cameron, and D.~Newth,
  ``Understanding human mobility from twitter,'' \emph{PloS one}, vol.~10,
  no.~7, p. e0131469, 2015.

\bibitem{khaidem2020optimizing}
L.~Khaidem, M.~Luca, F.~Yang, A.~Anand, B.~Lepri, and W.~Dong, ``Optimizing
  transportation dynamics at a city-scale using a reinforcement learning
  framework,'' \emph{IEEE Access}, vol.~8, pp. 171\,528--171\,541, 2020.

\bibitem{barbosa2018human}
H.~Barbosa, M.~Barthelemy, G.~Ghoshal, C.~R. James, M.~Lenormand, T.~Louail,
  R.~Menezes, J.~J. Ramasco, F.~Simini, and M.~Tomasini, ``Human mobility:
  Models and applications,'' \emph{Physics Reports}, vol. 734, pp. 1--74, 2018.

\bibitem{luca2022modeling}
M.~Luca, B.~Lepri, E.~Frias-Martinez, and A.~Lutu, ``Modeling international
  mobility using roaming cell phone traces during covid-19 pandemic,''
  \emph{EPJ Data Science}, vol.~11, no.~1, p.~22, 2022.

\bibitem{mejova2009sentiment}
Y.~Mejova, ``Sentiment analysis: An overview,'' \emph{University of Iowa,
  Computer Science Department}, 2009.

\bibitem{birjali2021comprehensive}
M.~Birjali, M.~Kasri, and A.~Beni-Hssane, ``A comprehensive survey on sentiment
  analysis: Approaches, challenges and trends,'' \emph{Knowledge-Based
  Systems}, vol. 226, p. 107134, 2021.

\bibitem{huber2007graphs}
W.~Huber, V.~J. Carey, L.~Long, S.~Falcon, and R.~Gentleman, ``Graphs in
  molecular biology,'' \emph{BMC bioinformatics}, vol.~8, no.~6, pp. 1--14,
  2007.

\bibitem{longa2022efficient}
A.~Longa, G.~Cencetti, B.~Lepri, and A.~Passerini, ``An efficient procedure for
  mining egocentric temporal motifs,'' \emph{Data Mining and Knowledge
  Discovery}, vol.~36, no.~1, pp. 355--378, 2022.

\bibitem{longa2022neighbourhood}
A.~Longa, G.~Cencetti, S.~Lehmann, A.~Passerini, and B.~Lepri, ``Neighbourhood
  matching creates realistic surrogate temporal networks,'' \emph{arXiv
  preprint arXiv:2205.08820}, 2022.

\bibitem{barnes1969graph}
J.~A. Barnes, ``Graph theory and social networks: A technical comment on
  connectedness and connectivity,'' \emph{Sociology}, vol.~3, no.~2, pp.
  215--232, 1969.

\bibitem{brandes2013social}
U.~Brandes, L.~C. Freeman, and D.~Wagner, ``Social networks,'' 2013.

\bibitem{samoaa2022tep}
H.~P. Samoaa, A.~Longa, M.~Mohamad, M.~H. Chehreghani, and P.~Leitner,
  ``Tep-gnn: Accurate execution time prediction of functional tests using graph
  neural networks,'' in \emph{Product-Focused Software Process Improvement:
  23rd International Conference, PROFES 2022, Jyv{\"a}skyl{\"a}, Finland,
  November 21--23, 2022, Proceedings}.\hskip 1em plus 0.5em minus 0.4em\relax
  Springer, 2022, pp. 464--479.

\bibitem{samoaa2023unified}
P.~Samoaa, L.~Aronsson, A.~Longa, P.~Leitner, and M.~H. Chehreghani, ``A
  unified active learning framework for annotating graph data with application
  to software source code performance prediction,'' \emph{arXiv preprint
  arXiv:2304.13032}, 2023.

\bibitem{ficara2021criminal}
A.~Ficara, L.~Cavallaro, F.~Curreri, G.~Fiumara, P.~De~Meo, O.~Bagdasar,
  W.~Song, and A.~Liotta, ``Criminal networks analysis in missing data
  scenarios through graph distances,'' \emph{PLoS one}, vol.~16, no.~8, p.
  e0255067, 2021.

\bibitem{luca2021leveraging}
M.~Luca, G.~Barlacchi, N.~Oliver, and B.~Lepri, ``Leveraging mobile phone data
  for migration flows,'' \emph{arXiv preprint arXiv:2105.14956}, 2021.

\bibitem{nastase2015survey}
V.~Nastase, R.~Mihalcea, and D.~R. Radev, ``A survey of graphs in natural
  language processing,'' \emph{Natural Language Engineering}, vol.~21, no.~5,
  pp. 665--698, 2015.

\bibitem{yang2017tensor}
Y.~Yang, D.~Krompass, and V.~Tresp, ``Tensor-train recurrent neural networks
  for video classification,'' in \emph{International Conference on Machine
  Learning}.\hskip 1em plus 0.5em minus 0.4em\relax PMLR, 2017, pp. 3891--3900.

\bibitem{novikov2015tensorizing}
A.~Novikov, D.~Podoprikhin, A.~Osokin, and D.~P. Vetrov, ``Tensorizing neural
  networks,'' \emph{Advances in neural information processing systems},
  vol.~28, 2015.

\bibitem{taboada2011lexicon}
M.~Taboada, J.~Brooke, M.~Tofiloski, K.~Voll, and M.~Stede, ``Lexicon-based
  methods for sentiment analysis,'' \emph{Computational linguistics}, vol.~37,
  no.~2, pp. 267--307, 2011.

\bibitem{le2015twitter}
B.~Le and H.~Nguyen, ``Twitter sentiment analysis using machine learning
  techniques,'' in \emph{Advanced Computational Methods for Knowledge
  Engineering}.\hskip 1em plus 0.5em minus 0.4em\relax Springer, 2015, pp.
  279--289.

\bibitem{zainuddin2018hybrid}
N.~Zainuddin, A.~Selamat, and R.~Ibrahim, ``Hybrid sentiment classification on
  twitter aspect-based sentiment analysis,'' \emph{Applied Intelligence},
  vol.~48, no.~5, pp. 1218--1232, 2018.

\bibitem{habimana2020sentiment}
O.~Habimana, Y.~Li, R.~Li, X.~Gu, and G.~Yu, ``Sentiment analysis using deep
  learning approaches: an overview,'' \emph{Science China Information
  Sciences}, vol.~63, no.~1, pp. 1--36, 2020.

\bibitem{vaswani2017attention}
A.~Vaswani, N.~Shazeer, N.~Parmar, J.~Uszkoreit, L.~Jones, A.~N. Gomez,
  {\L}.~Kaiser, and I.~Polosukhin, ``Attention is all you need,''
  \emph{Advances in neural information processing systems}, vol.~30, 2017.

\bibitem{lin2021survey}
T.~Lin, Y.~Wang, X.~Liu, and X.~Qiu, ``A survey of transformers,'' \emph{arXiv
  preprint arXiv:2106.04554}, 2021.

\bibitem{khan2021transformers}
S.~Khan, M.~Naseer, M.~Hayat, S.~W. Zamir, F.~S. Khan, and M.~Shah,
  ``Transformers in vision: A survey,'' \emph{ACM Computing Surveys (CSUR)},
  2021.

\bibitem{devlin2018bert}
J.~Devlin, M.-W. Chang, K.~Lee, and K.~Toutanova, ``Bert: Pre-training of deep
  bidirectional transformers for language understanding,'' \emph{arXiv preprint
  arXiv:1810.04805}, 2018.

\bibitem{xu2019bert}
H.~Xu, B.~Liu, L.~Shu, and P.~S. Yu, ``Bert post-training for review reading
  comprehension and aspect-based sentiment analysis,'' \emph{arXiv preprint
  arXiv:1904.02232}, 2019.

\bibitem{sun2019utilizing}
C.~Sun, L.~Huang, and X.~Qiu, ``Utilizing bert for aspect-based sentiment
  analysis via constructing auxiliary sentence,'' \emph{arXiv preprint
  arXiv:1903.09588}, 2019.

\bibitem{hoang2019aspect}
M.~Hoang, O.~A. Bihorac, and J.~Rouces, ``Aspect-based sentiment analysis using
  bert,'' in \emph{Proceedings of the 22nd nordic conference on computational
  linguistics}, 2019, pp. 187--196.

\bibitem{li2019exploiting}
X.~Li, L.~Bing, W.~Zhang, and W.~Lam, ``Exploiting bert for end-to-end
  aspect-based sentiment analysis,'' \emph{arXiv preprint arXiv:1910.00883},
  2019.

\bibitem{pota2020effective}
M.~Pota, M.~Ventura, R.~Catelli, and M.~Esposito, ``An effective bert-based
  pipeline for twitter sentiment analysis: A case study in italian,''
  \emph{Sensors}, vol.~21, no.~1, p. 133, 2020.

\bibitem{nguyen2022emotion}
A.~Nguyen, A.~Longa, M.~Luca, J.~Kaul, and G.~Lopez, ``Emotion analysis using
  multilayered networks for graphical representation of tweets,'' \emph{IEEE
  Access}, vol.~10, pp. 99\,467--99\,478, 2022.

\bibitem{Bishop:1995:NNP:525960}
C.~M. Bishop, \emph{Neural Networks for Pattern Recognition}.\hskip 1em plus
  0.5em minus 0.4em\relax New York, NY, USA: Oxford University Press, Inc.,
  1995.

\bibitem{ekman1992there}
P.~Ekman, ``Are there basic emotions?'' 1992.

\bibitem{neumann2012priming}
R.~Neumann and L.~Lozo, ``Priming the activation of fear and disgust: evidence
  for semantic processing.'' \emph{Emotion}, vol.~12, no.~2, p. 223, 2012.

\bibitem{edwards2006experimental}
S.~Edwards and P.~M. Salkovskis, ``An experimental demonstration that fear, but
  not disgust, is associated with return of fear in phobias,'' \emph{Journal of
  Anxiety Disorders}, vol.~20, no.~1, pp. 58--71, 2006.

\bibitem{govindarajan2020help}
V.~S. Govindarajan, B.~T. Chen, R.~Warholic, K.~Erk, and J.~J. Li, ``Help! need
  advice on identifying advice,'' \emph{arXiv preprint arXiv:2010.02494}, 2020.

\bibitem{conneau2019unsupervised}
A.~Conneau, K.~Khandelwal, N.~Goyal, V.~Chaudhary, G.~Wenzek, F.~Guzm{\'a}n,
  E.~Grave, M.~Ott, L.~Zettlemoyer, and V.~Stoyanov, ``Unsupervised
  cross-lingual representation learning at scale,'' \emph{arXiv preprint
  arXiv:1911.02116}, 2019.

\bibitem{devlin2018open}
J.~Devlin and M.-W. Chang, ``Open sourcing bert: State-of-the-art pre-training
  for natural language processing,'' \emph{Google AI Blog}, vol.~2, 2018.

\bibitem{lee2020patent}
J.-S. Lee and J.~Hsiang, ``Patent classification by fine-tuning bert language
  model,'' \emph{World Patent Information}, vol.~61, p. 101965, 2020.

\bibitem{chen2021leveraging}
K.~Chen, B.~Cosgro, O.~Domfeh, A.~Stern, G.~Korkmaz, and N.~A. Kattampallil,
  ``Leveraging google bert to detect and measure innovation discussed in news
  articles,'' in \emph{2021 Systems and Information Engineering Design
  Symposium (SIEDS)}.\hskip 1em plus 0.5em minus 0.4em\relax IEEE, 2021, pp.
  1--6.

\bibitem{sousa2019bert}
M.~G. Sousa, K.~Sakiyama, L.~de~Souza~Rodrigues, P.~H. Moraes, E.~R. Fernandes,
  and E.~T. Matsubara, ``Bert for stock market sentiment analysis,'' in
  \emph{2019 IEEE 31st International Conference on Tools with Artificial
  Intelligence (ICTAI)}.\hskip 1em plus 0.5em minus 0.4em\relax IEEE, 2019, pp.
  1597--1601.

\bibitem{yao2019graph}
L.~Yao, C.~Mao, and Y.~Luo, ``Graph convolutional networks for text
  classification,'' in \emph{Proceedings of the AAAI conference on artificial
  intelligence}, vol.~33, no.~01, 2019, pp. 7370--7377.

\end{thebibliography}

\begin{IEEEbiography}[{\includegraphics[width=1in,height=1.25in,clip,keepaspectratio]{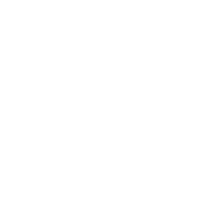}}]{Gabriel Lopez} is a Computer Scientist working at Pulse.io. He has worked in the past in Senior and Managerial positions in Machine Learning. He is broadly interested in all aspects of predicting human behavior and complex systems. His research activities involve developing new systems to reduce the unpredictability of the non-deterministic systems using neural networks.
\end{IEEEbiography}

\begin{IEEEbiography}[{\includegraphics[width=1in,height=1.25in,clip,keepaspectratio]{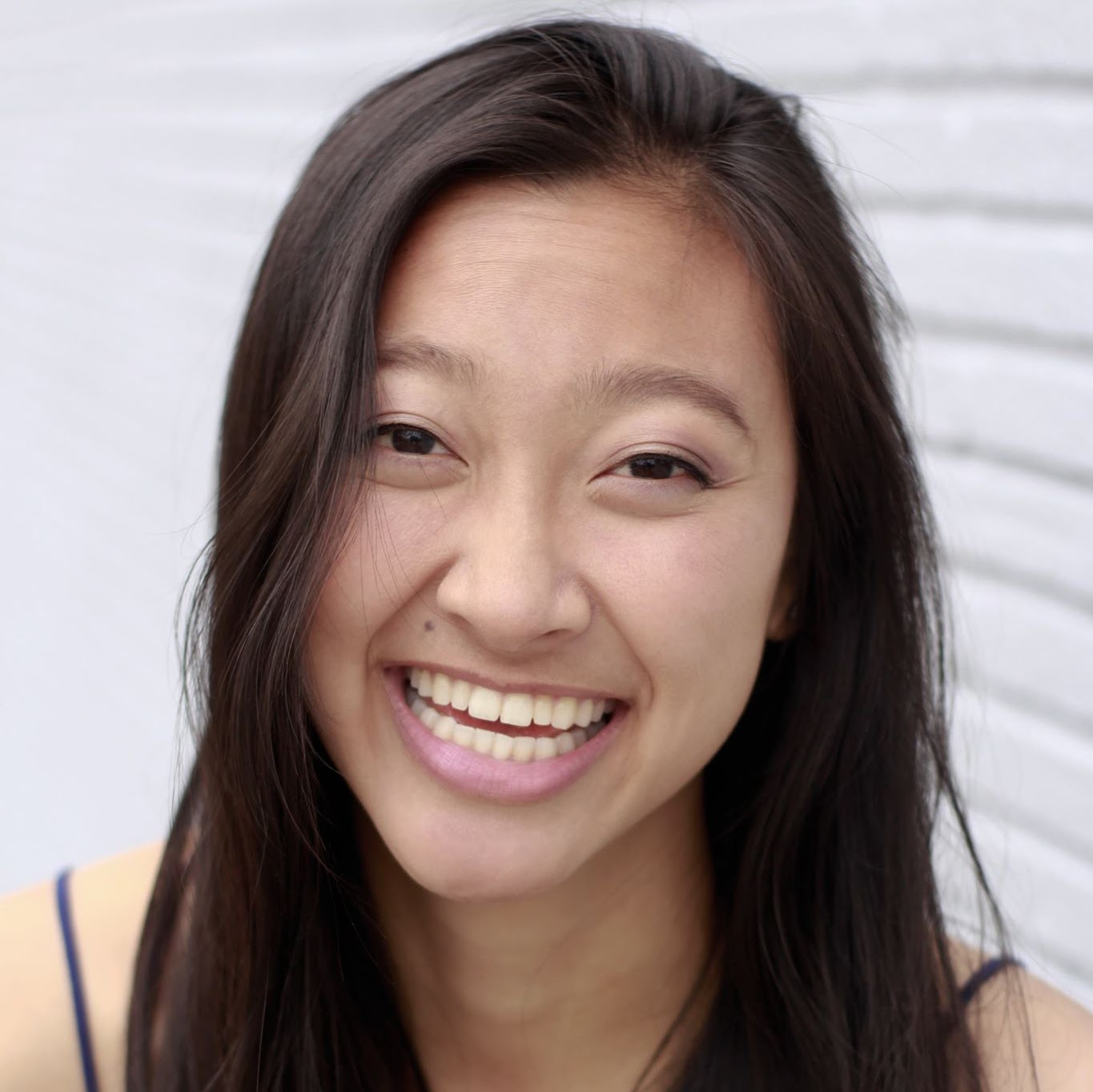}}]{Anna Nguyen} is consulting with Pulse.io. She has an MSc in Data Science from Birkbeck University and an MSc in Psychological Sciences from University College London (UCL). Professional portfolio includes leading projects that utilize natural language processing (NLP), graph modeling, and back-end development using Django. Previous research focuses on Generative Adversarial Networks (GANs) and their expansion of use to commercial industries. 
\end{IEEEbiography}

\begin{IEEEbiography}[{\includegraphics[width=1in,height=1.25in,clip,keepaspectratio]{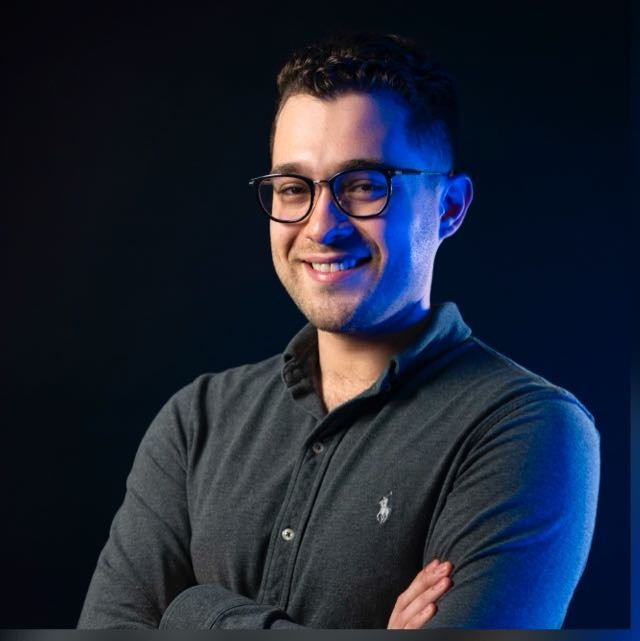}}]{Joe Kaul} is a marketing professional with almost a decade of experience in senior management in the industry. Kaul founded Pulse in 2020 with the intention of providing clearer insights into emotional intent of consumers by leveraging big data insights. Kaul has scaled his marketing agency group to a multi-million revenue group, with specific expertise in scaling return on investment by enhancing customer journey and conveying messaging in a compelling, emotionally evocative way.
\end{IEEEbiography}

\EOD
\end{document}